\documentclass[10pt,onecolumn,letterpaper]{article}
\usepackage{cvpr}
\usepackage{eso-pic}
\usepackage{times}
\usepackage{epsfig}
\usepackage{graphicx}
\usepackage{amsthm}
\usepackage{amsmath}
\usepackage{amssymb}
\usepackage[breaklinks=true,bookmarks=false]{hyperref}

\numberwithin{equation}{subsection}
\usepackage{datetime}
\usepackage[a4paper,margin=1in,headheight=15pt,headsep=24pt]{geometry}

\usepackage{fancyhdr}

\usepackage{algorithm}
\usepackage{algorithmic}

\usepackage{multirow}
\usepackage{color} 
\usepackage{enumitem}

\usepackage{algorithm}
\usepackage{algorithmic}

\usepackage{amsmath}
\usepackage{amssymb}
\usepackage{mathtools}
\usepackage{amsthm}
\usepackage{annotate-equations}
\usepackage{multirow} 
\usepackage{booktabs} 
\usepackage{colortbl}
\usepackage{makecell}

\cvprfinalcopy 
\def\cvprPaperID{****} 
\allowdisplaybreaks

\usepackage{bibentry}

\begin{document}
\lhead{}
\lfoot{\date{\today},\date{\currenttime}}
\rfoot{NGD for DL}

\title{Remodeling Semantic Correlations in Vision-Language Fine-Tuning}
\author{
\textbf{Xiangyang Wu}$^{1}$\quad
\textbf{Liu Liu}$^{2,1*}$\quad\\
\textbf{Baosheng Yu}$^{3}$\quad
\textbf{Jiayan Qiu}$^{4}$\quad
\textbf{Zhenwei Shi}$^{3}$\quad
\\
[0.5em]
$^1$Hangzhou International Innovation Institute, Beihang University\\
$^2$School of Artificial Intelligence, Beihang University\\
$^3$Nanyang Technological University \quad
$^4$University of Leicester\\
$^5$School of Astronautics, Beihang University\\
}
\maketitle
\begingroup
\renewcommand\thefootnote{*}
\footnotetext{Corresponding author: \texttt{liuliubh@buaa.edu.cn}}
\endgroup

\begin{abstract}
Vision-language fine-tuning has emerged as an efficient paradigm for constructing multimodal foundation models. While textual context often highlights semantic relationships within an image, existing fine-tuning methods typically overlook this information when aligning vision and language, thus leading to suboptimal performance. 
    Toward solving this problem, we propose a method that can improve multimodal alignment and fusion based on both semantics and relationships.
    Specifically, we first extract multilevel semantic features from different vision encoder to capture more visual cues of the relationships. 
    Then, we learn to project the vision features to group related semantics, among which are more likely to have relationships. 
    Finally, we fuse the visual features with the textual by using inheritable cross-attention, where we globally remove the redundant visual relationships by discarding visual-language feature pairs with low correlation. 
    We evaluate our proposed method on eight foundation models and two downstream tasks, visual question answering and image captioning, and show that it  outperforms all existing methods. Codes are publicly accessible at \url{https://github.com/simonmonmonmonn/LSRM}.
\end{abstract}


\section{Introduction}

Vision-language models (VLMs), capable of jointly processing visual and linguistic information, have demonstrated remarkable performance on multimodal downstream tasks, thanks to their powerful cross-modal comprehension capabilities. In recent years, advancements in large-scale pretrained models~\cite{tang2024rethinking} for natural language processing~\cite{achiam2023gpt, touvron2023llama} and computer vision \cite{radford2021learning} has led to the rise of parameter-efficient fine-tuning (PEFT) techniques, which have become the core paradigm for efficiently constructing multimodal systems. 

\begin{figure}[t]
    \centering
    \includegraphics[width=0.55\linewidth]{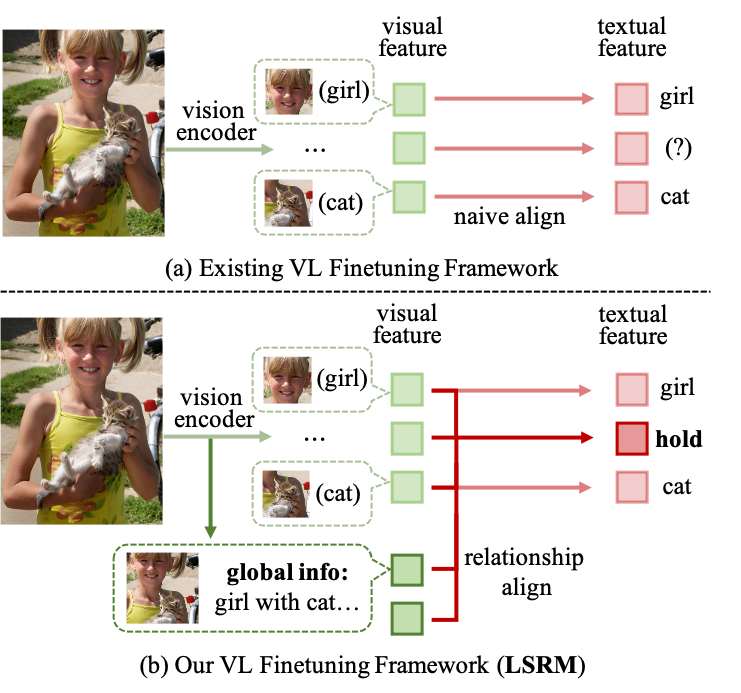}
    \caption{Overview of our paper: (a) Existing VL fine-tuning methods have weak ability to capture semantic relationships; 
    (b) Our method extracts global information from vision encoder and strengthens relationship modeling through better alignment.
    }
    \label{fig:abs}
\end{figure}

The construction of multimodal models comprises three stages: vision encoding, cross-modal alignment, and vision-language fusion. In the vision encoding stage, a pretrained vision encoder transforms input images into structured semantic features. The cross-modal alignment stage bridges the semantic gap between the two modalities by projecting visual features into the language model’s representation space through a trainable alignment module. 
Typical implementations of this stage include linear projection layers~\cite{liu2024improved, chen2023minigpt}, or other customized projection methods~\cite{mckinzie2024mm1, ye2023mplug, bai2023qwen}. 
Finally, the vision-language fusion stage facilitates cross-modal interaction mechanisms for joint multimodal semantic reasoning. In the multimodal scenario, PEFT aims to introduce efficient alignment and fusion modules to bridge the vision encoder and language model.

Recent studies reveal that while cross-attention-based fine-tuning methods have achieved remarkable progress in improving the inference efficiency of multimodal models~\cite{jie2024memory}, they still lack thorough research to bridge the semantic gap between vision models and vision-language models. This fundamental limitation stems from the pretraining objectives of vision encoders: existing approaches predominantly optimize them for single-semantic classification tasks rather than modeling inter-semantic relationships. For instance, in the classic multimodal framework CLIP \cite{radford2021learning}, the training objective of its visual encoder exhibits clear classification properties: it requires the encoder to output categories matching a series of natural language representations.  This representational deficiency directly conflicts with the core requirement of vision-language models: the ability to parse cross-semantic relationships based on textual prompts. For example, when describing an image of ``a girl holding a cat", the model must not only recognize the independent semantics of ``girl" and ``cat", but also comprehend the relationship denoted by “holding.” The semantic gap between these two model paradigms remains a critical bottleneck constraining the performance of multimodal systems. Existing architectures lack specialized designs for multimodal semantic relationship modeling: in vision-encoding stage, the adopted final-layer visual features with classification properties are deficient in complex semantic relationship information; in the alignment and fusion stages, the simple MLP projector and naive cross-attention fusion method lack the ability to remodel semantic relationships.

We propose the learnable semantic relationship method (LSRM), a cross-attention-based fine-tuning framework that systematically enhances the understanding of multimodal semantic relationships through. Firstly, we adopt the multilevel information fusion method for image encoding, extracting multilevel image features from different layers of the encoder. Considering that intermediate-layer outputs can better preserve semantic relationships while final-layer features focus on high-level semantic abstraction, this strategy will balance local correlations and global semantics. Secondly, we employ the semantic relationship projector to align multilevel image features to the textual space. We observe that dimensionality reduction matrices in conventional projectors implicitly perform semantic grouping through their sparse structure. Our semantic relationship projector addresses this by introducing a learnable diagonal matrix $\mathbf{\Lambda}$ after activation to dynamically amplify critical semantic relationship groups while suppressing irrelevant signals. Finally, we use inheritable cross-attention to fuse the aligned image features with text features from language model. Considering that the correlation strength between text and image tokens is an inherent property, we share this connection in the multi-layer cross-attention fusion with a weight matrix $\mathbf{M}$ shared between layers. This matrix progressively attenuates attention responses to redundant token pairs across network depths while preserving high-confidence correlations, enabling hierarchical refinement from local semantic interactions to globally consistent multimodal understanding through persistent focus on strongly correlated cross-modal patterns.

To evaluate the LSRM framework, we conduct extensive experiments on two downstream tasks—visual question answering and image captioning—using eight foundation models of varying scales from the LLaMA1-LLaMA3 \cite{grattafiori2024llama,touvron2023llama,touvron2023llama2} and Vicuna \cite{chiang2023vicuna} series. On the ScienceQA visual question answering benchmark \cite{lu2022learn}, after 20 epochs of fine-tuning with the LLaMA-7B language model and CLIP ViT-L/14 vision encoder, our method achieves 93.94\% accuracy on the test set, surpassing the state-of-the-art cross-attention fine-tuning framework MemVP \cite{jie2024memory} by 0.87\%. Consistent performance advantages are observed across all seven other foundation models. For image captioning, our framework achieves performance comparable to the current best methods on both BLEU-4 \cite{papineni2002bleu} and CIDEr \cite{vedantam2015cider} metrics, further verifying its robustness.

Our main contributions are summarized as follows:
\begin{itemize}
    \item We propose the \textbf{L}earnable \textbf{S}emantic \textbf{R}elationship \textbf{M}ethod (\textbf{LSRM}) to enhance vision-language semantic relationship modeling;
    
    \item We introduce the multilevel information fusion strategy, which fully leverages the advantages of both semantic representations by aggregation of intermediate and final layer outputs from the vision encoder;
    
    \item We develop the semantic relationship projector, achieving adaptive enhancement of key semantic groups through a learnable diagonal weight matrix;
    
    \item We introduce the inheritable cross-attention mechanism, which ensures sustained focus on strongly correlated cross-modal interactions through an inheritable weight matrix.
\end{itemize}

\section{Related works}

\subsection{Parameter-Efficient Fine-Tuning}

Compared to traditional fine-tuning methods, PEFT significantly reduces the number of parameters to be updated. PEFT methods can be categorized based on their impact on the model:
1) Methods acting directly on the language model, which adjust model parameters or structure to modify the knowledge stored in the model. Key methods include BitFit \cite{zaken2021bitfit}, Adapter \cite{houlsby2019parameter}, and LoRA \cite{hu2021lora}. BitFit trains only the bias values in pre-trained models, minimizing trainable parameters.
2) Methods acting on language model inputs or intermediate-layer inputs, such as Prefix-Tuning \cite{li2021prefix} and Prompt-Tuning \cite{lester2021power}, which modify input representations. 

Many PEFT techniques designed for language models can be applied to multimodal architectures. For instance, Adversarial DuAl Prompt Tuning (ADAPT) \cite{10901852} achieves efficient Unsupervised Domain Adaptation (UDA) through domain alignment via adversarial fine-tuning of both textual and visual prompts. MetaPrompt \cite{10431687} enhances the model's domain generalization capability through a dual-modality prompt tuning network and an alternating episodic training algorithm. CLIP4STR \cite{10816351} significantly boosts \textit{image-text alignment} and irregular text recognition capabilities with its dual-branch model architecture and triple optimization strategy. MemVP \cite{jie2024memory} tailors PEFT for vision-language models by training lightweight alignment modules and incorporating cross-attention for information fusion in the language model’s feed-forward layers.

\subsection{Vision-Language Models}

In multimodal tasks with pre-trained vision encoders and large language models, visual and textual representations reside in distinct embedding spaces due to modality-specific pretraining. Aligning these representations is crucial for effective fusion. 
A common approach is input-space prompting-based fine-tuning, where visual inputs are projected and concatenated with text inputs. 
Representative methods include VL-Adapter \cite{sung2022vl}, VL-PET \cite{hu2023vl}, LaVIN \cite{luo2024cheap}, and LLaVA \cite{liu2024visual}, though this increases input sequence lengths, hindering contextual associations and raising inference latency. Cross-attention-based alignment, proposed by Flamingo \cite{alayrac2022flamingo} and BLIP \cite{li2022blip, li2023blip}, projects visual data into keys and values for cross-attention, achieving fusion without increasing input length. UniAdapter \cite{lu2023uniadapter} and MemVP \cite{jie2024memory} further optimize this with frame-aware attention and simplified cross-attention computations, improving inference speed and efficiency. Despite progress, parameter-efficient methods for enhancing cross-attention with better semantic relationship understanding remain underexplored. Our work aims to develop more efficient, lightweight, and generalizable fine-tuning frameworks.

\section{Method}\label{Method-0}
\begin{figure}[t]
    \centering
    \includegraphics[width=1.0\linewidth]{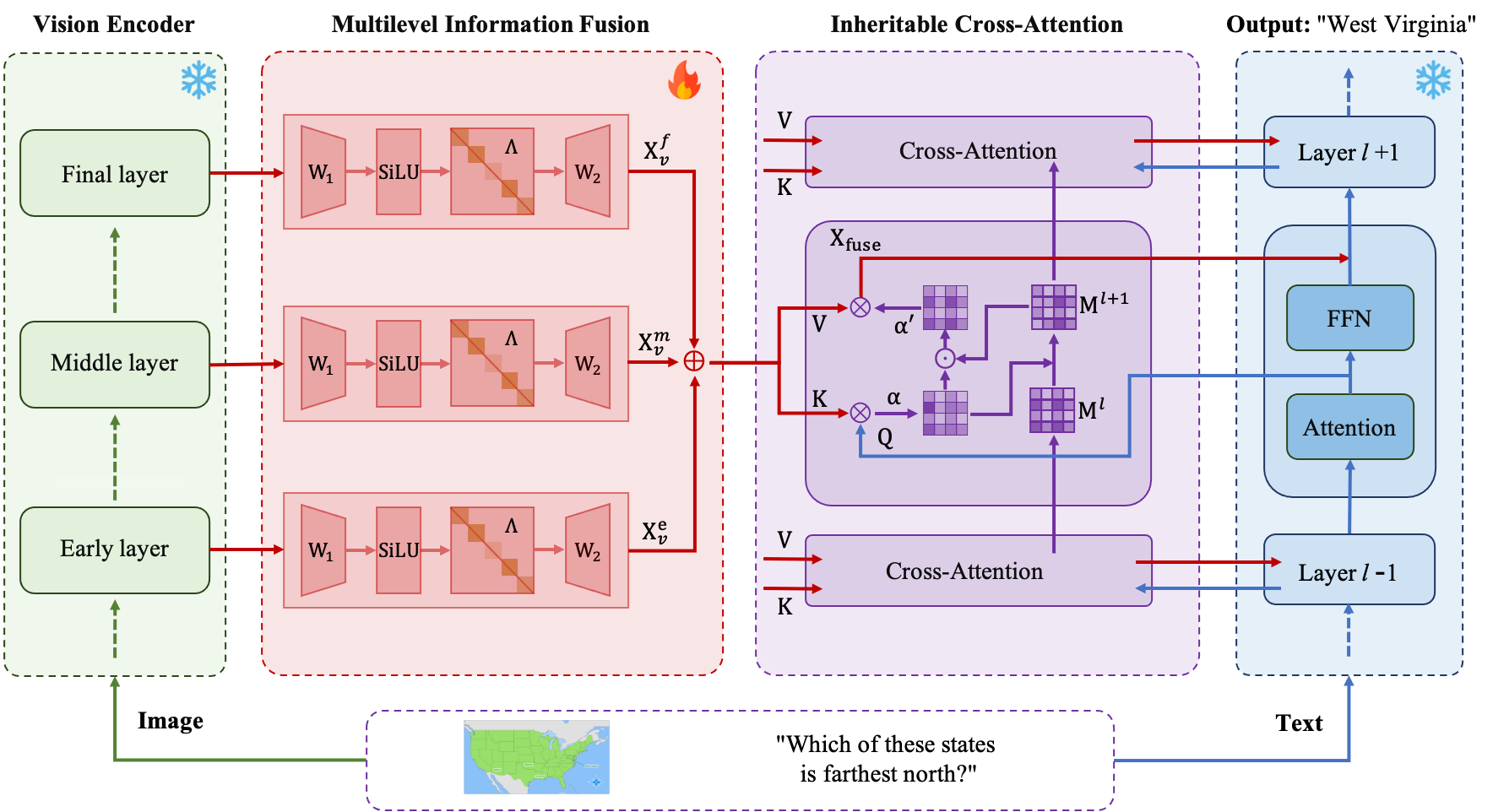}
    \caption{The main LSRM(Learnable Semantic Relationship Method) framework. ``$\oplus$" denotes matrix addition, ``$\otimes$" denotes matrix multiplication, and ``$\odot$" denotes the Hadamard product of matrices (i.e., element-wise multiplication). 
    The example of input and output are from ScienceQA \cite{lu2022learn} datasets}.
    \label{fig:framework}
\end{figure}

In this section, we begin by introducing the multilevel information fusion strategy during the vision encoding stage. Next, we present the semantic relationship projector for cross-modal alignment. Finally, we elaborate on the inheritable cross-attention mechanism for vision-language fusion. Figure~\ref{fig:framework} illustrates the overall LSRM framework for efficient vision-language fine-tuning. We also provide the pseudocode algorithm of LSRM in Algorithm \ref{Alg:newFramework}.

\subsection{Multilevel Information Fusion}\label{Method-1}

 Vision encoders are typically optimized for image recognition and classification tasks during the pre-training, resulting in final-layer outputs that capture high-level visual representations but struggle with semantic relationships. In contrast, intermediate layers retain semantic relationship information yet produce low-level representations due to incomplete processing. To address this, this module aims to combine the strengths of both: preserving the semantic expressiveness of final-layer outputs while leveraging the relational information in intermediate layers.

Our framework employs a multilevel information fusion method as follows. Given the significant divergence in feature spaces across different outputs, we independently train a visual projector for each selected feature layer. The final multilevel information fusion adopts an efficient averaging strategy, 

\begin{align*}
    \mathbf{X}_{v} = \frac{1}{L}\big[\text{Proj}_f(\mathbf{X}_{v}^{f})+\text{Proj}_m(\mathbf{X}_{v}^{m}) + \text{Proj}_e(\mathbf{X}_{v}^{e})+\cdots\big],
\end{align*}  

where \( L \) denotes the number of selected feature layers (including intermediate and final layers), \( \mathbf{X}_{v}^{f}, \mathbf{X}_{v}^{m}, \mathbf{X}_{v}^{e} \) represents the vision encoder output of the final, intermediate and earlier layer, and \( \text{Proj} \) refers to the mutually independent projector for multimodal alignment, which will be elaborated in Section~\ref{Method-2}.  

\subsection{Semantic Relationship Projector}
\label{Method-2}

Due to the difference in representation spaces between vision encoders and language models, encoded visual features must undergo cross-modal alignment through a trainable projector before participating in subsequent multimodal interactions. Under the parameter-efficient fine-tuning paradigm, conventional projectors typically adopts a dimension reduction-activation-dimension lifting architecture. For visual features with dimension \(d_1\) and linguistic features with dimension \(d_2\), a projector with hidden dimension \(d_h\) (\(d_h \ll d_1, d_2\)) is: 
\begin{equation}  
    \text{Proj}(\mathbf{X}_v) = \lambda \cdot \text{SiLU}(\mathbf{X}_v \mathbf{W}_1) \mathbf{W}_2,  
\end{equation}  
where \(\mathbf{W}_1 \in \mathbb{R}^{d_1 \times d_h}\) is the dimension reduction matrix, \(\mathbf{W}_2 \in \mathbb{R}^{d_h \times d_2}\) is the dimension lifting matrix, \(\text{SiLU}\) denotes the sigmoid linear unit activation function \cite{elfwing2018sigmoid}, defined as \(\text{SiLU}(x) = x \cdot \text{Sigmoid}(x)\), \(\lambda\) is a scale hyperparameter set manually.

The sparsity of \(\mathbf{W}_1\) implicitly performs semantic grouping: interpreting \(\mathbf{X}_v\) as a multi-category probability distribution, this operation constructs cross-semantic relationship groups. However, in the dense features generated after nonlinear activation, different semantic relationship groups exhibit varying importance. Traditional architectures lack dynamic weight adjustment capabilities. To address this, we propose the semantic relationship projector (SRProj) by introducing a learnable diagonal weight matrix \( \mathbf{\Lambda} = \text{diag}(\lambda_1,...,\lambda_{d_h})\) after the activation layer to adaptively enhance critical semantic relationships:  
\begin{align}  
    \text{SRProj}(\mathbf{X}_v) = (\mathbf{\Lambda} \cdot \text{SiLU}(\mathbf{X}_v \mathbf{W}_1)) \mathbf{W}_2.  
\end{align} 
Building upon the multilevel information fusion method proposed in Section~\ref{Method-1}, the overall workflow can be formally expressed as:  
\begin{align}
    \nonumber
    \mathbf{X}_{v} = \frac{1}{L}\big[&\text{SRProj}_f(\mathbf{X}_{v}^{f})+\text{SRProj}_m(\mathbf{X}_{v}^{m})\\
    +\ &\text{SRProj}_e(\mathbf{X}_{v}^{e})+\cdots\big].
\end{align}  
To avoid adding excessive model complexity, we ultimately use only a single intermediate-layer feature. In this configuration, the specialized expression formulated as:  
\begin{align}  
    \mathbf{X}_v = \frac{1}{2}[\text{SRProj}_m(\mathbf{X}_v^{m}) + \text{SRProj}_f(\mathbf{X}_v^{f}) ],  
\end{align}  
where \(\mathbf{X}_v^{m}\) and \(\mathbf{X}_v^{f}\) denote the intermediate-layer and final-layer output features, respectively.  

\begin{algorithm}[t]
\caption{LSRM Framework}
\label{Alg:newFramework}
\begin{algorithmic}[1]  
    \STATE \textbf{Input:} text $x_\text{tex}$, image $x_\text{img}$,
    \STATE \textbf{Output:} multimodal feature $\mathbf{X}_t$
    \STATE $\mathbf{X}_t \xleftarrow{} \text{LLM-embedding}(x_\text{tex})$
    \STATE $\triangleright$ $\textbf{Multilevel Information Fusion}$

    \STATE $\mathbf{X}_v^{f}, \mathbf{X}_v^{m} \xleftarrow{} \text{VisionEncoder}(x_\text{img})$ 

    \STATE $\triangleright$ $\textbf{Semantic Relationship Projector}$
    \STATE $\mathbf{X}_v \xleftarrow{} \frac{1}{2}\left[\text{SRProj}_m(\mathbf{X}_v^{m}) + \text{SRProj}_f(\mathbf{X}_v^{f}) \right]  $

    \STATE \textbf{Initialize:} $\mathbf{M} = \textbf{I}_{M \times N}$
    
    \FOR{Layer \textbf{in} LLM}
        \STATE $\mathbf{X}_t \xleftarrow{} \text{Layer.Attn}(\mathbf{X}_t)$
        \STATE $\mathbf{Q} \xleftarrow{} \mathbf{X}_t, \mathbf{K} \xleftarrow{}  \mathbf{X}_v + \mathbf{P}_1, \mathbf{V} \xleftarrow{}  \mathbf{X}_v + \mathbf{P}_2$
        \STATE $\triangleright \ \textbf{Inheritable Cross-Attention}$
        \STATE $\mathbf{\alpha} \xleftarrow{} \text{SiLU}(\mathbf{Q} \mathbf{K}^T)$
        \STATE 
        $
                \mathbf{M}_{ij} \xleftarrow{} 
                \begin{cases}  
                \lambda \mathbf{M}_{ij} & \text{if } \mathbf{\alpha}_{ij} \in \text{lowest } \delta\% \text{ of } \mathbf{\alpha}_{i} \\  
                \mathbf{M}_{ij} & \text{otherwise}.  
                \end{cases}         
        $
        \STATE $\mathbf{\alpha{'}} \xleftarrow{} \mathbf{\alpha} \cdot \mathbf{M}$
        \STATE $\mathbf{X}_\text{fuse} \xleftarrow{} \mathbf{\alpha{'}} \cdot \mathbf{V}$
        \STATE $\mathbf{X}_t \xleftarrow{} \mathbf{X}_\text{fuse} + \text{Layer.FFN}(\mathbf{X}_t)$
        
    \ENDFOR
    
\end{algorithmic}
\end{algorithm}

\subsection{Inheritable Cross-Attention}
\label{Method-3}

In this paper, we follow the cross-attention computation framework in MemVP \cite{jie2024memory}, and the expression is as follows:
\begin{align}  
    \mathbf{X}_\text{fuse}=\text{Cross-Attn}(\mathbf{Q},\mathbf{K},\mathbf{V}) = \text{SiLU}\left(\mathbf{Q}\mathbf{K}^\top\right)\mathbf{V}.
\end{align} 
Concretely, queries (\(\mathbf{Q}\)), keys (\(\mathbf{K}\)) and values (\(\mathbf{V}\)) are formulated as follows:

\begin{align}  
    \mathbf{Q} = \mathbf{X}_t,  
    \mathbf{K} = \mathbf{X}_v + \mathbf{P}_1,  
    \mathbf{V} = \mathbf{X}_v + \mathbf{P}_2,  
\end{align}  

where $\mathbf{X}_\text{fuse}$ means multimodal fused output of cross-attention. \(\mathbf{X}_t \) denotes the intermediate-layer output of the language model, and \(\mathbf{P}_1, \mathbf{P}_2\) are two trainable positional embedding matrices. 

The attention score computation (\textit{i.e.}, $\mathbf{\alpha} = \text{SiLU}(\mathbf{Q} \mathbf{K}^\top)$) constitutes the core process for establishing vision-language cross-modal semantic correlations. Each value in the $\mathbf{\alpha}$ matrix quantifies the association strength between a visual token and a text token. Low-correlation noisy token pairs often introduce interference through spontaneous allocation of attention scores. Suppressing these low-confidence attention responses can implicitly enhance the interaction weights of highly correlated semantic units.  

Notably, when cross-attention modules are embedded within each Transformer layer of the visual model, the learned cross-modal correlation patterns exhibit hierarchical transfer properties. We propose an inheritable cross-attention module for inter-layer correlation knowledge sharing: by inheriting effective semantic correlation information learned in shallow layers to deeper networks, this approach suppresses attention scores of redundant token pairs, thereby guiding the model to persistently focus on strongly correlated cross-modal interactions. To achieve this, we introduce an inheritable weight matrix $\mathbf{M}$ to accumulate the suppression weights of token pairs. Initialized as an all-ones matrix, this mechanism performs weight decay on the lowest $\delta\%$ token pairs in each row (corresponding to each text token) of the attention scores $\mathbf{\alpha}_{ij}$ during each cross-attention computation as:  
\begin{align}  
    \mathbf{M}_{ij}^{l} = \begin{cases}  \lambda \mathbf{M}_{ij}^{l-1} & \text{if } \mathbf{\alpha}_{ij}^{l} \in \text{lowest } \delta\% \text{ of } \mathbf{\alpha}_{i}^{l}, \\  \mathbf{M}_{ij}^{l-1} & \text{otherwise}.  \end{cases}  
\end{align}  
Here, the script $l$ denotes the cross-attention module inserted at the $l$-th layer of the language model, $i, j$ means the $i$th and the $j$th token of text token and vision token, and $\lambda \in (0,1)$ is a preset decay factor.  
Building upon the weight matrix $\mathbf{M}$, the Inheritable Cross-Attention is calculated as follows:  
\begin{align}  
    \text{Cross-Attn}(\mathbf{Q}, \mathbf{K}, \mathbf{V}) =  \left[ \mathbf{M} \odot \text{SiLU}\left( \mathbf{Q}\mathbf{K}^\top \right) \right] \mathbf{V} ,
\end{align}  
where $\odot$ denotes the Hadamard product of matrices (i.e., element-wise multiplication).

Given that inheritable cross-attention serves as an enhancement mechanism for cross-attention and should be introduced only after the model has developed foundational attention modeling capabilities, we introduce the hyperparameter \texttt{shift\_epoch} to delay the activation of inheritable cross-attention until the training process has reached a specified number of epochs.

\begin{table*}[!t]
    \renewcommand\tabcolsep{3pt}
    \centering
    \small
    \begin{tabular}{lcc|ccc|ccc|cc|c}
        \toprule
        \multirow{2}{*}{Method}                          & \multicolumn{2}{c|}{\#Param} & \multicolumn{3}{c|}{Subject} & \multicolumn{3}{c|}{Context Modality} & \multicolumn{2}{c|}{Grade} & \multirow{2}{*}{Average}                                                                                                       \\
                                                         & Trainable & LLM & NAT            & SOC            & LAN            & TXT            & IMG            & NO             & G1-6           & G7-12          &                \\
        \midrule
        \multicolumn{2}{l}{\it{Zero-/few-shot methods}}&&&&&&&&&&                                                                                                                                                                \\
        Human~\cite{lu2022learn}                         & -         & -   & 90.23          & 84.97          & 87.48          & 89.60          & 87.50          & 88.10          & 91.59          & 82.42          & 88.40          \\
        GPT-4~\cite{achiam2023gpt}                       & -         & -   & 84.06          & 73.45          & 87.36          & 81.87          & 70.75          & 90.73          & 84.69          & 79.10          & 82.69          \\
        \midrule
        \multicolumn{2}{l}{\it Full training methods}&&&&&&&&&&                                                                                                                                                                            \\
        UnifiedQA~\cite{lu2022learn}                     & 223M      & -   & 71.00          & 76.04          & 78.91          & 66.42          & 66.53          & 81.81          & 77.06          & 68.82          & 74.11          \\
        MM-CoT$_\text{Base}$~\cite{zhang2023multimodal}  & 223M      & -   & 87.52          & 77.17          & 85.82          & 87.88          & 82.90          & 86.83          & 84.65          & 85.37          & 84.91          \\
        LLaVA~\cite{liu2024visual}                       & 7B        & 7B  & -              & -              & -              & -              & -              & -              & -              & -              & 89.84          \\
        CoMD (Vicuna-7B)\cite{lu2022learn}                & 7B        & 7B  &91.83	        &95.95	         &88.91   	      &90.91	       &89.94	        &91.08	         &92.47	          &90.97	       &91.94           \\
        \midrule
        \multicolumn{2}{l}{\it{PEFT methods with LLaMA-7B}}&&&&&&&&&&                                                                                                                                                               \\
        PILL (LLaMA-7B)\cite{yin2024pill}              &45M 	     &7B  &90.36	        &95.84	         &89.27	          &89.39	       &88.65	        &91.71	         &92.11	          &89.65	       &91.23           \\
        LLaVA-LoRA~\cite{jie2024memory}                  & 4.4M      & 7B  & 91.70          & 94.60          & 86.09          & 91.25          & 90.28          & 88.64          & 91.52          & 89.65          & 90.85          \\
        LLaMA-Adapter~\cite{zhang2023llama}              & 1.8M      & 7B  & 84.37          & 88.30          & 84.36          & 83.72          & 80.32          & 86.90          & 85.83          & 84.05          & 85.19          \\
        MemVP~\cite{jie2024memory}                       & 3.9M      & 7B  & 94.45          & 95.05          & 88.64          & 93.99          & 92.36          & 90.94          & 93.10          & 93.01          & 93.07          \\
        \rowcolor[gray]{0.9}
        LSRM (ours) & 3.9M      & 7B  &95.16           & 95.28          & 90.36          & 94.87          & 93.41          & 92.20          & 94.20          & 93.47          & \textbf{93.94 }         \\
        \midrule
        \multicolumn{2}{l}{\it{PEFT methods with LLaMA-13B}}&&&&&&&&&&                                                                                                                                                                 \\
        LaVIN (LLaMA-13B)~\cite{luo2024cheap}                        & 5.4M      & 13B & 90.32          & 94.38          & 87.73          & 89.44          & 87.65          & 90.31          & 91.19          & 89.26          & 90.50          \\
        MemVP~\cite{jie2024memory}                       & 5.5M      & 13B & 95.07          & 95.15          & 90.00          & 94.43          & 92.86          & 92.47          & 93.61          & \textbf{94.07}          & 93.78          \\
        \rowcolor[gray]{0.9}
        LSRM ({ours})                                       & 5.5M      & 13B & \textbf{96.09} & \textbf{95.61}          & \textbf{90.00} & \textbf{95.55} & \textbf{94.00} & \textbf{92.47} & \textbf{94.68} & 93.94 & \textbf{94.41} \\
        \bottomrule
    \end{tabular}    
    \caption{Evaluation results on ScienceQA test set with LLaMA-7B and LLaMA-13B. NAT = natural science, SOC = social science, LAN = language science, TXT = text context, IMG = image context, NO = no context, G1-6 = grades 1-6, G7-12 = grades 7-12.}
    \label{tab:llama}
\end{table*}

\begin{table*}
    \renewcommand\tabcolsep{3pt}
    \centering
    \small
    \begin{tabular}{lcc|ccc|ccc|cc|c}
        \toprule
        \multirow{2}{*}{Method}                          & \multicolumn{2}{c|}{\#Param} & \multicolumn{3}{c|}{Subject} & \multicolumn{3}{c|}{Context Modality} & \multicolumn{2}{c|}{Grade} & \multirow{2}{*}{Average}                                                                                                       \\
                                                         & Trainable & LLM & NAT            & SOC            & LAN            & TXT            & IMG            & NO             & G1-6           & G7-12          &                \\
        \midrule
        \multicolumn{2}{l}{\emph{PEFT methods with Vicuna-7B}}&&&&&&&&&&                                                                                                                                                                          \\
        LaVIN (Vicuna)~\cite{luo2024cheap}                        & 3.8M      & 7B & 89.25          & 94.94          & 85.24          & 88.51          & 87.46          & 88.08          & 90.16          & 88.07          & 89.41          \\
        MemVP~\cite{jie2024memory}                       & 3.9M      & 7B & 93.92          & 94.94          & \textbf{89.00}          & 93.65          & 91.87          & \textbf{91.50}          & 92.88          & 92.81          & 92.86          \\
        
        \rowcolor[gray]{0.9}
        LSRM (ours)                                       & 3.9M      & 7B & \textbf{94.94} & \textbf{95.39}          & 88.73 & \textbf{94.83} & \textbf{93.21} &       90.80 & \textbf{93.61} & \textbf{93.08} & \textbf{93.42} \\
        \midrule
        \multicolumn{2}{l}{\emph{PEFT methods with LLaMA2-7B}}&&&&&&&&&&                                                                                                                                                                          \\
        MemVP~\cite{jie2024memory}                       & 3.9M      & 7B & 95.07          & 95.50          & 88.27          & 94.28          & 92.76          & 91.29          & 93.72          & 92.81          & 93.40          \\
        \rowcolor[gray]{0.9}
         LSRM (ours)                                         & 3.9M      & 7B & \textbf{95.34} & \textbf{95.95}          & \textbf{89.55} & \textbf{95.11} & \textbf{93.65} &       \textbf{91.71} & \textbf{94.02} & \textbf{93.87} & \textbf{93.96} \\
        \midrule
            \multicolumn{2}{l}{\emph{PEFT methods with LLaMA2-13B}}&&&&&&&&&&                                                                                                                                                                          \\
        MemVP~\cite{jie2024memory}                       & 5.5M      & 7B & 94.45          & 95.28          & 91.00          & 94.04          & 92.66          & 92.75          & 93.39          & 94.33          & 93.73          \\
        \rowcolor[gray]{0.9}
        LSRM (ours)                                     & 5.5M      & 13B & \textbf{95.60} & \textbf{95.28}          & \textbf{91.36} & \textbf{95.26} & \textbf{93.70} &       \textbf{93.31} & \textbf{94.38} & \textbf{94.53} & \textbf{94.44} \\
        \midrule
            \multicolumn{2}{l}{\emph{PEFT methods with LLaMA3}}&&&&&&&&&&                                                                                                                                                                          \\
        MemVP-1B~\cite{jie2024memory}                       & 2.3M      & 1B & 92.58          & 94.15          & 85.73          & 92.08          & 90.68          & 88.64          & 91.48          & 90.51          & 91.13          \\
        \rowcolor[gray]{0.9}
        LSRM-1B (ours)                                      & 2.3M      & 1B & 93.43 & 94.83          & 87.27 & 92.52 & 91.52 &       90.03 & 92.03 & 92.29 & 92.12 \\
        \rowcolor[gray]{0.9}
        LSRM-3B (ours)                                       & 3.0M      & 3B & 95.16& 95.16         & 90.09 & 94.92 & 93.41 &       92.06 & 93.72 & 94.07 & 93.85 \\
        \rowcolor[gray]{0.9}
        LSRM-8B (ours)                                       & 3.9M      & 8B & \textbf{95.78} & \textbf{95.61}          & \textbf{92.73} & \textbf{95.16} & \textbf{93.55} &       \textbf{94.84} & \textbf{95.01} & \textbf{94.86} & \textbf{94.95} \\
        \midrule
    \end{tabular}
    \caption{Evaluation results on ScienceQA test set with Vicuna, LLaMA2 and LLaMA3. NAT = natural science, SOC = social science, LAN = language science, TXT = text context, IMG = image context, NO = no context, G1-6 = grades 1-6, G7-12 = grades 7-12.}
    \label{tab:llama2}
\end{table*}

\section{Experiments}

In this section, we evaluate the LSRM framework through experiments on diverse datasets and models. Section~\ref{settings} outlines the setup, followed by a performance comparison on VQA and image captioning tasks in Section~\ref{Quantitative}. Ablation studies and qualitative validations are shown in Section~\ref{ablation} and  Section~\ref{Qualitative}, respectively.

\subsection{Experimental Setups}\label{settings}

The experiments are primarily conducted on the ScienceQA \cite{lu2022learn} dataset for visual question answering.
Models are fine-tuned on the training set, with the average accuracy on the test set as the core evaluation metric. Additionally, for the image captioning task, we evaluate our method on the COCO Captions dataset \cite{chen2015microsoft} using the Karpathy split \cite{karpathy2015deep}, comparing results with existing methods through BLEU-4 and CIDEr scores. We validate the framework using a combination of open-source pretrained vision encoders and language models. For ScienceQA, we test 8 language models: LLaMA-7B/13B \cite{touvron2023llama}, LLaMA2-7B/13B \cite{touvron2023llama2}, LLaMA3-1B/3B/8B \cite{grattafiori2024llama}, and Vicuna-7B \cite{chiang2023vicuna}. For the COCO image captioning task, we use the LLaMA-13B model. Visual features in all experiments are extracted using the CLIP ViT-L/14 encoder \cite{radford2021learning}. 

For ScienceQA experiments, we follow MemVP's optimization settings. Specifically, fine-tuning is performed for 20 epochs on LLaMA-7B and LLaMA-13B models with a global batch size of 32. The initial learning rate is set to $9e^{-3}$ and decays via a cosine schedule. The multilevel information fusion module uses the intermediate-layer output (layer 12 out of 24 layers) from the vision encoder, with the fusion weight hyperparameter $s$ fixed at 0.1. The hidden dimension of the semantic relationship projector is set to 64, maintaining half the parameters of MemVP's projector to ensure consistency in total parameters. The inheritable cross-attention module uses hyperparameters $\delta=0.3$ and $\lambda=0.85$, with a training mode switch at epoch 14. More detailed hyperparameter configurations will be presented in appendix.

\begin{table}
    \renewcommand\tabcolsep{3pt}
    \centering
    \small
    \begin{tabular}{lc|cc}
        \toprule
        Method                                   & \#T. & BLEU-4 & CIDEr \\
        \midrule
        VisionLLM-H~\cite{wang2024visionllm}     & -    & 32.1   & 114.2 \\
        BLIP~\cite{li2022blip}                   & 583M & 40.4   & 136.7 \\
        BLIP-2~\cite{li2023blip}                 & 188M & 43.7   & 145.3 \\
        $^*$LLaMA-Adapter~\cite{gao2023llama} & 14M  & 36.2   & 122.2 \\
        $^*$MemVP~\cite{jie2024memory}            & 5.5M & 36.6   & 121.6 \\
        \rowcolor[gray]{0.9}
        $^*$LSRM~                              & 5.5M & 37.3   & 123.9 \\
        \bottomrule
    \end{tabular}
    \caption{Evaluation results on COCO caption using the Karpathy test split with LLaMA-13B as the language model. \#T. = trainable parameters. *PEFT methods.}
    \label{tab:coco}
\end{table}

\begin{table}[t]
    \renewcommand\tabcolsep{3pt}
    \centering
    \small
    \begin{tabular}{lc|cc}
        \toprule
        Method&\makecell[c]{\#Trainable\\ Params}&\makecell[c]{Training \\Time (s/batch)}&\makecell[c]{Inference \\Time (s/batch)}\\
        \midrule
         LLaVA-LoRA \footnotesize{7B}& 4.4M& 0.49 & 3.42 \\
         LaVIN \footnotesize{7B}& 3.8M& 0.39 & 2.06 \\
        MemVP \footnotesize{7B}& 3.9M& 0.28 & 1.88 \\
        MemVP \footnotesize{13B}& 5.5M& 0.46 & 3.07 \\
        \rowcolor[gray]{0.9}
        LSRM \footnotesize{7B}& 3.9M& 0.29 & 1.93 \\
        \rowcolor[gray]{0.9}
        LSRM \footnotesize{13B}& 5.5M& 0.48	& 3.17 \\
        \bottomrule
    \end{tabular}
    \caption{Measured on 8$\times$A800 GPUs without memory-saving or speed-up techniques (\emph{e.g.,} flash attention). The per-GPU batch size is 4 for training and 64 for inference.}
    \label{tab:efficiency}
\end{table}

\subsection{Quantitative Results}\label{Quantitative}

\subsubsection{Results on ScienceQA.}
We first validate the performance advantages of our model on the ScienceQA dataset, with results summarized in Table~\ref{tab:llama}. To comprehensively evaluate the effectiveness, we covers three representative categories of methods from the ScienceQA official benchmark \cite{lu2022learn}: zero-/few-shot learning, full training, and parameter-efficient fine-tuning (PEFT) methods. Experiments demonstrate that our method achieves significant performance superiority over all zero-/few-shot and full training baselines. In comparisons with PEFT methods, our model surpasses all baselines in terms of average accuracy. Specifically, fine-tuning experiments on LLaMA-7B achieve a 0.87\% improvement over the current state-of-the-art method MemVP \cite{jie2024memory}, while the LLaMA-13B version exceeding previous best by 0.63\%. These performance improvements are achieved with only a minimal increase in model parameters.

We further evaluate the framework's performance on the LLaMA2 and LLaMA3 model families, with results shown in Table~\ref{tab:llama2}. Experimental results demonstrate that our framework consistently outperforms existing fine-tuning frameworks across 7 pretrained models with different parameter scales. Notably, the LLaMA3-1B based implementation achieves the most significant improvement, showing a 0.99\% test accuracy gain over existing fine-tuning frameworks. This advantage likely stems from our method's prioritized optimization of semantic relationship learning capabilities, which effectively enhances cross-modal modeling capacities of smaller-scale models.

\subsubsection{Results on COCO Caption.}
We also validate the performance of the LSRM framework on the COCO Captions dataset. The experiments employ the baseline results of the MemVP framework for comparative analysis. As shown in Table~\ref{tab:coco}, despite having significantly fewer trainable parameters than full-parameter fine-tuning methods \cite{li2022blip, li2023blip}, our framework exhibits no substantial performance gap in testing. Specifically, LSRM outperforms the MemVP baseline by 0.9 points on BLEU-4 and 2.3 points on CIDEr. These results further verify the comprehensive performance advantages of our method under parameter-efficient constraints.

\begin{table}
    \renewcommand\tabcolsep{3pt}
    \centering
    \small
    \begin{tabular}{lc|c}
        \toprule
        Setting&\makecell[c]{\#Trainable\\ Params}&\makecell[c]{Average \\Accuracy}\\
        \midrule
        Baseline& 3.9M& 92.78\\
        + Multilevel Information Fusion& 3.9M& 93.47\\
        + Semantic Relationship Projector& 3.9M& 93.75\\
        + Inheritable Cross-Attention& 3.9M& \textbf{93.94}\\
        \bottomrule
    \end{tabular}
    \caption{Ablation study of each module in our  framework with LLaMA-7B as the language model and CLIP as the vision encoder.}
    \label{tab:ablation}
\end{table}

\subsubsection{Efficiency.} To validate the computational performance of our proposed method relative to existing approaches, we evaluate the efficiency of LSRM on ScienceQA+LLaMA-7B/13B tasks using 8$\times$A800 GPUs for both training and inference. Experimental results are presented in Table~\ref{tab:efficiency}. 
Compared to conventional input-space prompt-based fine-tuning methods (e.g., LLaVA-LoRA, LaVIN), LSRM demonstrates significant advantages in training and inference speed. When compared with the state-of-the-art cross-attention-based method MemVP, LSRM exhibits slightly increased speeds on the 7B model. Nevertheless, LSRM shows no significant disadvantage in computational efficiency compared to SOTA methods.

\subsection{Ablation Studies} \label{ablation}

To verify the effectiveness of each module, we conduct the ablation study of the main framework by employing the LLaMA-7B model and the ScienceQA dataset. The framework comprises three core components: multilevel information fusion, semantic relationship projector, and inheritable cross-attention. As shown in Table~\ref{tab:ablation}, ablation experiments start with a baseline model that integrates none of these components, achieving a accuracy of 92.78\%. Upon introducing multilevel information fusion, the accuracy improves to 93.47\%. Further integrating the semantic relationship projector increases the accuracy to 93.75\%, and finally, incorporating the inheritable cross-attention module elevates the accuracy to 93.94\%. The results progressively validate the independent effectiveness of each component and confirm the effect of component collaboration on overall performance.
More ablation experiments regarding hyperparameters and architectural details will be presented in appendix.

\subsection{Qualitative Results} \label{Qualitative}

To validate the theoretical analyses in Section~\ref{Method-3} about inheritable cross-attention, we conduct qualitative visualization experiments using the LLaMA-13B-based model on the COCO Captions dataset. Given the inherent divergence between the model's semantic processing and human cognitive patterns, our analysis focuses on representative samples with explicitly interpretable decision rationales.

\begin{figure}[t]
    \centering
    \includegraphics[width=0.55\linewidth]{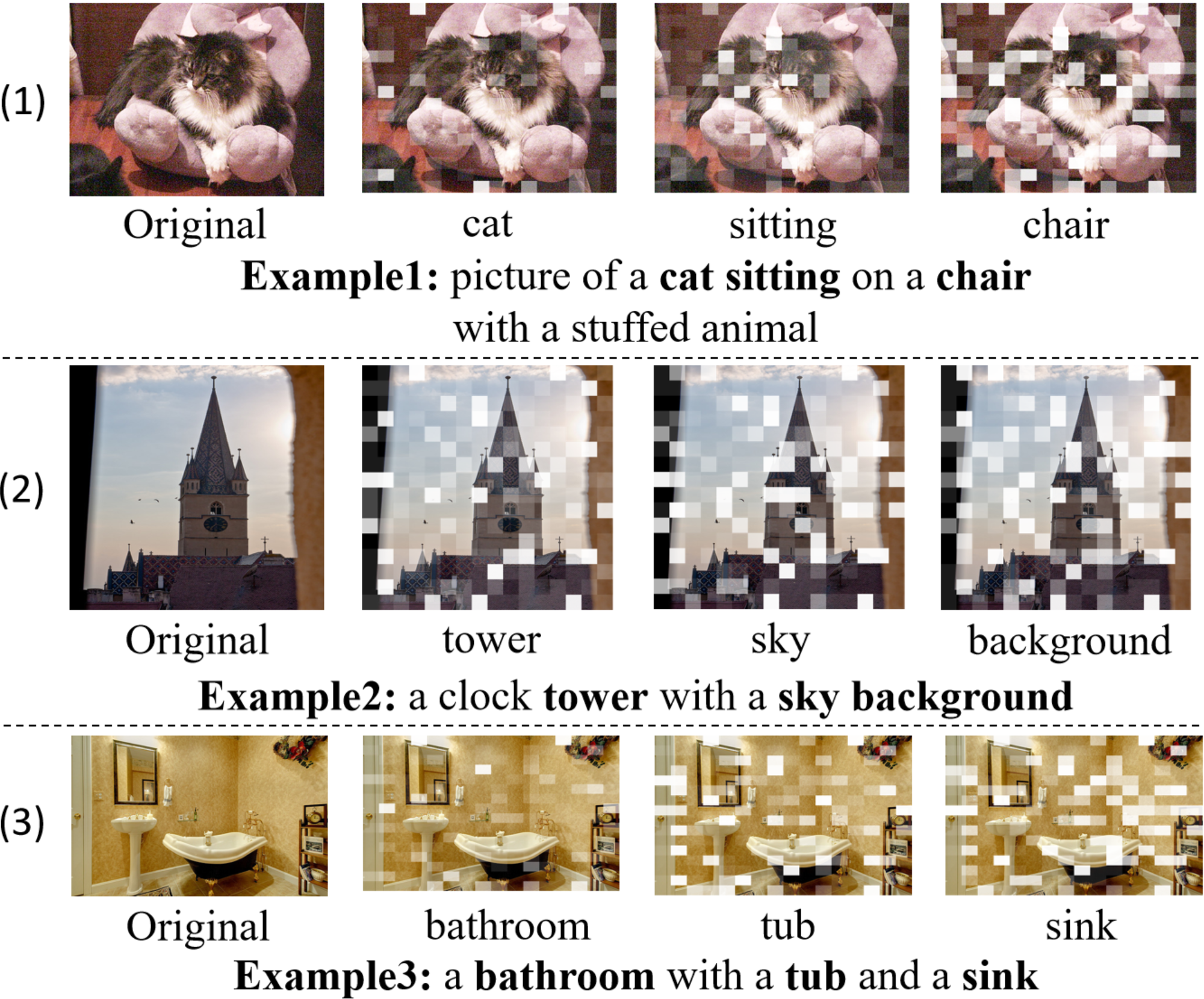}
    \caption{Visualization of Inheritable Cross-Attention.In each row, the left figure is the original image, while the middle and right figures demonstrate the value of inheritable matrix between two representative text tokens and each image tokens.}
    \label{fig:qualitative1}
\end{figure}

The core mechanism of the inheritable cross-attention module is implemented through the inheritable matrix $\mathbf{M}$. To analyze its functionality, we visualize the weight values of $\mathbf{M}$ by mapping the attention weights between specific text tokens and image regions to transparency levels at corresponding image locations, where lower weights correspond to higher transparency, as shown in Figure~\ref{fig:qualitative1}.
Three key patterns are revealed: 
firstly, low-weight regions (high transparency) generally exhibit weak semantic relationship with the current text token;
secondly, when a visual entity (\textit{e.g.}, the tower structure in Example2) is mentioned in preceding descriptions, its attention weights are significantly reduced during subsequent text generation, achieving the automatic suppression of historical information;
thirdly, when generating global text descriptions, the range of the image attended to by matrix $\mathbf{M}$ also expands to capture global semantics. 
For instance, in Example3, the image range attended to when describing the scene ``bathroom" is significantly larger than that when describing the local semantics ``tub" and ``sink".
Notably, some low-weight regions still maintain strong relevance in human cognition, suggesting that the model’s attention mechanism does not fully capture the semantic importance assigned by humans.
We hypothesize that this discrepancy originates from fundamental differences in perceptual mechanisms between the model and humans.

\section{Conclusion}

In this paper, we tackle the challenge of improving semantic relationship understanding in vision-language fine-tuning. We introduce LSRM, a learnable semantic relationship method comprising three integrated components: multilevel information fusion for remodeling semantic relationships, a semantic relationship projector to enhance key semantic groups, and inheritable cross-attention to establish inter-layer attention suppression. Extensive experiments across eight baseline models and two downstream tasks demonstrate the effectiveness of the proposed method, emphasizing the critical role of remodeling semantic relationships in vision-language fine-tuning.

{\small
 \bibliographystyle{unsrt}
 \bibliography{egbib}
}

\appendix
\onecolumn
\section{Appendix}

\subsection{Experiment Details} \label{config}

\subsubsection{Experiments on ScienceQA}
For the LLaMA-based fine-tuning experiments on the ScienceQA dataset, supplementary hyperparameter settings are documented in Table~\ref{tab:hyper-llama}. To ensure equitable comparison with existing methods, our hyperparameter configurations strictly follow the implementation protocol of MemVP.

\begin{table}[h]  
\centering  
\begin{tabular}{lcll}  
\toprule
Method&position embedding&projector &vision adapter\\\midrule
MemVP (7B)& $len = 320$&$h_d = 128$ &$h_d = 12$\\
MemVP (13B)&$len = 400$&$h_d = 128$ &$h_d = 12$\\
\rowcolor{lightgray}LSRM (7B)&$len = 320$&$h_d = 64$(each) &$h_d = 12$\\
\rowcolor{lightgray}LSRM(13B)&$len = 400$&$h_d = 64$(each) &$h_d = 12$\\
\bottomrule
\end{tabular}  
\caption{\textbf{More Hyperparameters on LLaMA.}} 
\label{tab:hyper-llama}  
\end{table} 

The hyperparameter configurations for fine-tuning experiments based on LLaMA2/3 and Vicuna are documented in Table~\ref{tab:hyper-llama2}. For better-performing pre-trained models (such as LLaMA3-8B), we slightly reduced the value of \(\lambda\) to balance the impact of increased model depth. Meanwhile, we advanced the timing of switching training modes because high-performance models are relatively easier to develop multimodal capabilities.

\begin{table}[h]
\centering  
\begin{tabular}{lclll}  
\toprule
Method&position embedding&projector   &vision adapter&Inheritable Cross-Attn\\\midrule
MemVP (LLaMA2-7B)& $len = 320$&$h_d = 128$   &$h_d = 12$&-\\
MemVP(Vicuna-7B)&$len = 320$&$h_d = 128$   &$h_d = 12$&-\\
MemVP(LLaMA2-13B)& $len = 400$& $h_d = 128$ &$h_d = 12$&-\\
MemVP(LLaMA3-1B)& $len = 320$& $h_d = 128$ & $h_d = 12$&-\\
\rowcolor{lightgray}LSRM (LLaMA2-7B)&$len = 320$&$h_d = 64$(each)   &$h_d = 12$&$\lambda = 0.85, \delta = 0.3, \text{shift epoch} = 14$\\
\rowcolor{lightgray}LSRM(LLaMA2-7B)&$len = 320$&$h_d = 64$(each)   &$h_d = 12$&$\lambda = 0.85, \delta = 0.3, \text{shift epoch} = 14$\\
\rowcolor{lightgray}LSRM(LLaMA2-13B)& $len = 400$& $h_d = 64$(each)   &$h_d = 12$&$\lambda = 0.85, \delta = 0.3, \text{shift epoch} = 9$\\
 \rowcolor{lightgray}LSRM(LLaMA3-1B)& $len = 320$& $h_d = 64$(each)   & $h_d = 12$&$\lambda = 0.85, \delta = 0.3, \text{shift epoch} = 14$\\
 \rowcolor{lightgray}LSRM(LLaMA3-3B)& $len = 320$& $h_d = 64$(each)   & $h_d = 12$&$\lambda = 0.85, \delta = 0.3, \text{shift epoch} = 14$\\
 \rowcolor{lightgray}LSRM(LLaMA3-8B)& $len = 320$& $h_d = 64$(each)   & $h_d = 12$&$\lambda = 0.9, \delta = 0.3, \text{shift epoch} = 9$\\
\bottomrule

\end{tabular}  
\caption{\textbf{Hyperparameters on LLaMA2/3 and Vicuna.}} 

\label{tab:hyper-llama2}  
\end{table} 	

\subsubsection{Experiment on COCO Caption}

For the LLaMA-based fine-tuning experiments on the COCO Caption dataset, supplementary hyperparameter settings are documented in Table~\ref{tab:hyper-coco}. In the image captioning task, we increased \(\lambda\) and decreased \(delta\)to ensure that the model can explore more diverse outputs.

\begin{table}[h]
\centering  
\begin{tabular}{llclll}  
\toprule
Method &epoch&position embedding&projector   &vision adapter&Inheritable Cross-Attn\\\midrule
MemVP(LLaMA2-13B) &5& $len = 400$& $h_d = 128$ &$h_d = 12$&-\\
\rowcolor{lightgray}LSRM(LLaMA2-13B) &5& $len = 400$& $h_d = 64$(each)   &$h_d = 12$&$\lambda = 0.95, \delta = 0.1, \text{shift epoch} = 3$\\
\bottomrule

\end{tabular}  
\caption{\textbf{Hyperparameters on COCO Caption.}} 

\label{tab:hyper-coco}  
\end{table} 	

\subsection{More ablation study} \label{appendixabla}

\begin{figure*}[h]
    \centering
    \includegraphics[width=1\linewidth]{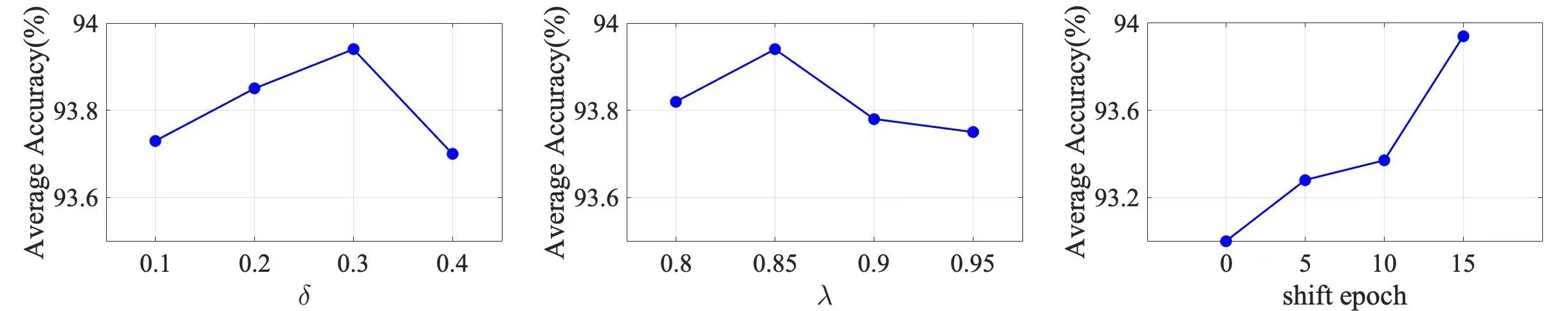}
    \caption{Comparison of different hyperparameter settings in the Inheritable Cross-Attention with LLaMA-7B as the language model.}
    \label{fig:hyper}
\end{figure*}

\subsubsection{Inheritable Cross-Attention.}
We investigate the performance sensitivity of hyperparameters $\delta$ and $\lambda$ in the inheritable cross-attention module through ablation experiments, and the results are  shown in Figure~\ref{fig:hyper}. Experiments demonstrate that when $\delta$ ranges from 0.1 to 0.4 and $\lambda$ ranges from 0.8 to 0.95, the model accuracy exhibits a unimodal distribution pattern, peaking at $\delta=0.3$ and $\lambda=0.85$ with 93.94\% accuracy. We hypothesize that excessively high $\delta$ or low $\lambda$ values overly restrict the model's capacity to explore weakly correlated semantics, whereas excessively low $\delta$ or high $\lambda$ values weaken the attention mechanism's suppression capability on irrelevant semantic associations. Furthermore, we validate the influence of training phase transition timing. Experimental data indicate that delaying the transition epoch leads to sustained growth in test accuracy. 
This phenomenon suggests that prematurely activating the attention inheritance mechanism during early training stages (when cross-modal alignment remains unstable) may disrupt the coordinated optimization process of vision-language semantic correlations.

\begin{table}[t]
\renewcommand{\arraystretch}{1.0}  
    \centering
    \small
    \begin{tabular}{cc|c}
        \toprule
        \#layers & Sources&  Average        \\
        \midrule
        2                     &$L, \frac{1}{2}L$                                       &\textbf{93.94} \\
        4                     &$L, \frac{3}{4}L, \frac{1}{2}L, \frac{1}{4}L$           &92.83 \\
        8                     &$L, \frac{7}{8}L, \cdots ,\frac{1}{4}L, \frac{1}{8}L$   &92.08 \\
        \bottomrule
    \end{tabular}
    \caption{Comparison of multilevel information from different intermediate layers.}
    \label{tab:ablation:num}
\end{table}

\subsubsection{Multilevel Information Fusion}
As discussed in Section 3.1, experimental results demonstrate that using single intermediate-layer features achieves optimal model performance. Table~\ref{tab:ablation:num} shows that increasing the number of selected features to 4 or 8 layers (including final-layer features) significantly degrades performance. We attribute this phenomenon to two factors: First, under parameter-efficiency constraints, multi-layer feature fusion reduces parameter allocation per visual projector, thereby weakening cross-modal alignment capability. Second, excessive intermediate-layer features dilute the contribution of final-layer features in fused representations, impairing precise semantic modeling capacity.
There are several methods for integrating low-level signals. We first designed an ablation study on the integration methods, including concatenation, addition, averaging, and weighted averaging. We report our test results in Table~\ref{tab:ablation:fuse}. From the results, the direct averaging method used in our framework proved to be the most effective.

The low-level visual signals can theoretically come from every layer of the vision encoder. To explore the impact of the source of the low-encoded visual signals, we designed corresponding ablation experiments and report the results in Table~\ref{tab:ablation:xfuse}. The results show that the closer the source of the low-encoded signal is to the middle layers of the vision encoder, the better the performance. Signals with lower levels are more difficult to extract effective information, while signals with higher encoding levels carry redundant information, both of which can lead to performance degradation.

\begin{table}[t]
    \centering
    \small
    \begin{tabular}{cc|c}
        \toprule
        Fusion Mode& proportion(high : low)& Average        \\
        \midrule
        None            &                       -& 92.78\\
        Concat&       -& 92.67\\
        Add&          -& 92.01\\
        Average&           (0.5, 0.5)& \textbf{93.94}\\
        Weighted Average&           (0.9, 0.1)& 92.67\\
        Weighted Average&           (0.6, 0.4)& 93.49\\
        Weighted Average&           (0.4, 0.6)& 93.09\\
        Weighted Average&           (0.1, 0.9)& 93.19\\
        \bottomrule
    \end{tabular}
    \caption{Comparison of different low-level information fusion mode with LLaMA-7B as the language model.}
    \label{tab:ablation:fuse}
\end{table}

\begin{table}
\renewcommand{\arraystretch}{1.3}  
    \centering
    \small
    \begin{tabular}{cc|c}
        \toprule
        Layer& Layer Number&  Average        \\
        \midrule
        $L - 1$&23       &93.02 \\
        $L - 2$&22       &93.02 \\
        $\frac{3}{4}L$&18          &93.42 \\
        $\frac{1}{2}L$&12          &\textbf{93.94} \\
        $\frac{1}{4}L$&8          &93.23 \\
        $0$(not-encoded)&0          &92.71 \\

        \bottomrule
    \end{tabular}
    \caption{Comparison of different sources of low-level information with LLaMA-7B as the language model.}
    \label{tab:ablation:xfuse}
\end{table}

\subsection{More qualitative result} \label{appendixqua}

\subsubsection{Multilevel Information Fusion}

\begin{figure}[t]
    \centering
    \includegraphics[width=1\linewidth]{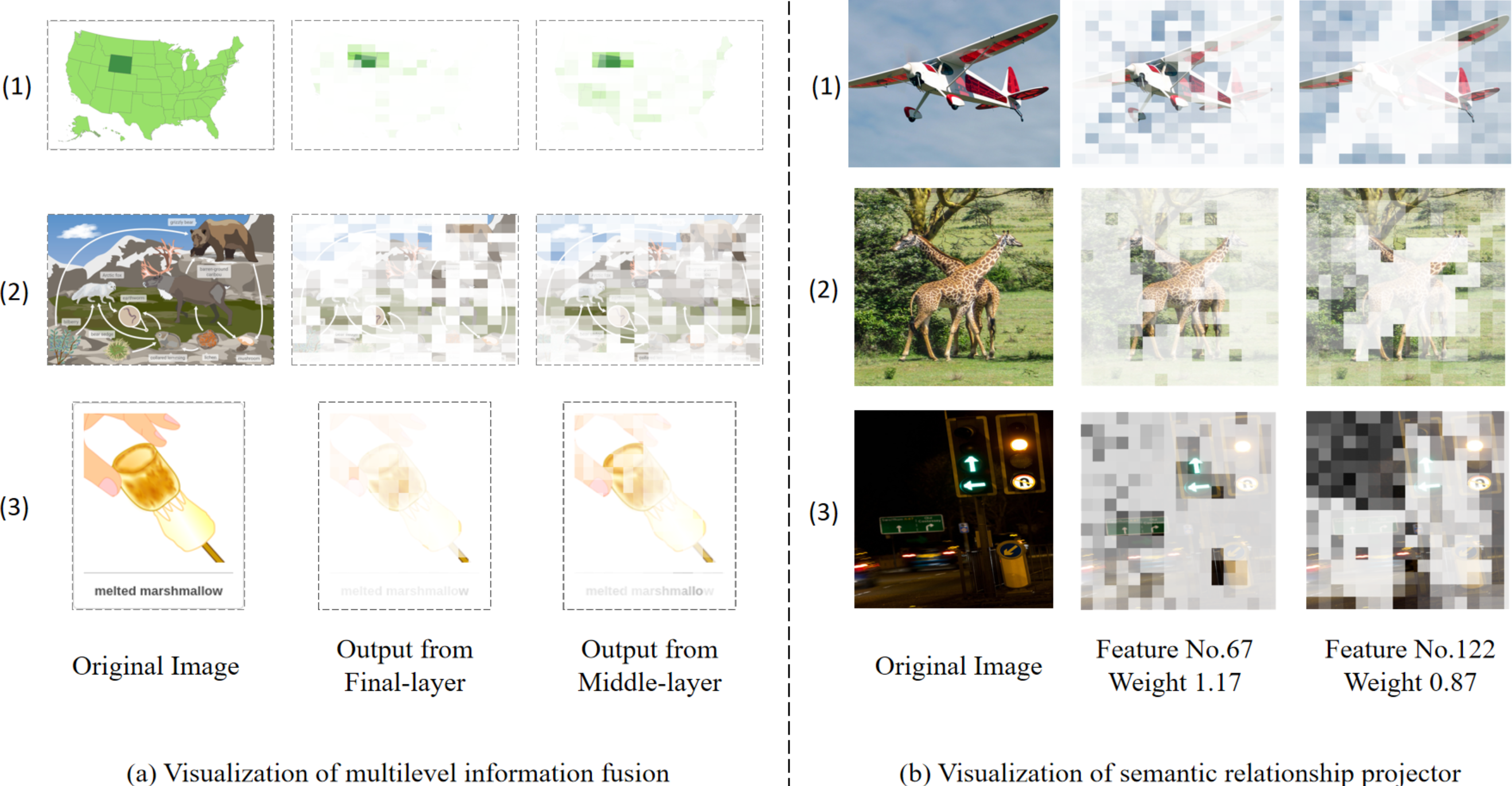}
    \caption{(a)Visualization of multilevel information fusion. In each row, the left figure is the original image, while the middle and right figures demonstrate the attention intensity (specifically, the maximum value of the projected output) for each image patch from the final layer output and intermediate layer output of the visual encoder, respectively. (b)Visualization of semantic relationship projector. In each row, the left figure is the original image, while the middle and right figures demonstrate the projection value of each image token across two hidden feature dimension of SRProj. ``Weight" denotes the corresponding value in $\Lambda$ of each dimension. ``Feature No." denotes the number of each feature dimension}
    \label{fig:qualitativeappendix}
\end{figure}

The advantage of multilevel information fusion over traditional methods lies in its incorporation of additional intermediate-layer information. To investigate the functional characteristics of this information, we conducted visualization experiments. In these experiments, we used the maximum value of the projected output vector per image patch as the metric for model attention intensity. We then visualized the outputs from both the final layer and intermediate layers of the visual encoder. The results reveal that the final layer output focuses on a significantly narrower range of image patches compared to the intermediate layer output. While this enables the final layer to more precisely localize critical image regions, it may also lead to the omission of certain important information. For instance, in sample group (3) in Figure~\ref{fig:qualitativeappendix}~(a), the final layer output clearly ignored the textual prompts at the image bottom, whereas the intermediate layer output exhibited markedly higher attention intensity towards this area.

\subsubsection{Semantic Relationship Projector.}
The core trainable parameter of the semantic relationship projector is the diagonal matrix $\mathbf{\Lambda}$ following the post-activation layer, which dynamically adjusts importance weights across feature dimensions, as analyzed in Section 3.2. Our visualization methodology maps the projection values of image tokens (i.e., spatial positions) to regional transparency levels, where regions with lower projection values exhibit higher transparency.
\begin{itemize}
    \item \textbf{Low-weight dimension (122nd)}: With $\lambda_{122}=0.87$, this dimension predominantly attends to semantically insignificant regions (\textit{e.g}., the sky background in the aircraft image shown in Figure~\ref{fig:qualitativeappendix}~(b)
    \item \textbf{High-weight dimension (67th)}: With $\lambda_{67}=1.17$, this dimension precisely focuses on critical semantic entities (\textit{e.g.}, the aircraft structure in Figure~\ref{fig:qualitativeappendix})
\end{itemize}

This phenomenon validates the adaptive adjustment mechanism of the $\mathbf{\Lambda}$ matrix for matrix for enhancing critical semantic features.

\end{document}